\documentclass{article}

\usepackage[final, nonatbib]{tackling_climate_workshop_style}

\usepackage[utf8]{inputenc} 
\usepackage[T1]{fontenc}    
\usepackage{hyperref}       
\usepackage{url}            
\usepackage{booktabs}       
\usepackage{amsfonts}       
\usepackage{nicefrac}       
\usepackage{microtype}      
\usepackage{graphicx}
\usepackage{biblatex}
\usepackage{wrapfig}
\usepackage{subcaption}
\title{Continuous Convolutional Neural Networks for Disruption Prediction in Nuclear Fusion Plasmas}

\author{%
  William F. Arnold \\
  Kim Jaechul School of AI \\
  KAIST \\
  \texttt{will@mli.kaist.ac.kr} \\
  \And
  Lucas Spangher \\
  Plasma Science and Fusion Center \\
  Massachussets Institute of Technology\\
  \texttt{spangher@psfc.mit.edu} \\
  \And 
  Cristina Rea\\
  Plasma Science and Fusion Center \\
  Massachussets Institute of Technology\\
}

\begin{document}

\maketitle

\begin{abstract}


Grid decarbonization for climate change requires dispatchable carbon-free energy like nuclear fusion. The tokamak concept offers a promising path for fusion, but one of the foremost challenges in implementation is the occurrence of energetic plasma disruptions. In this study, we delve into Machine Learning approaches to predict plasma state outcomes. Our contributions are twofold: (1) We present a novel application of Continuous Convolutional Neural Networks for disruption prediction and (2) We examine the advantages and disadvantages of continuous models over discrete models for disruption prediction by comparing our model with the previous, discrete state of the art, and show that continuous models offer significantly better performance (Area Under the Receiver Operating Characteristic Curve = 0.974 v.s. 0.799) with fewer parameters.

\end{abstract}

\section{Introduction}


Effectively combating climate change requires significant decarbonization of the grid. Nuclear fusion has long been considered a ``holy grail'' for producing on-demand energy without significant land demands or waste, and many have quantified beneficial carbon impacts \cite{GIULIANI2023113554, spangher2019characterizing}. Fusion was considered far away until recently, when renewed public-private enterprise in fusion has ushered in a wave of investment, interest, and supporting industries. 

One major approach to generating fusion in laboratory plasmas is \textit{magnetic confinement}, which leverages the tendency of plasmas to remain confined perpendicular to magnetic fields. Currently, the highest performing magnetic fusion concept is the \textit{tokamak}, a toroidal vacuum chamber with external magnetic coils that help generate a large plasma current. 

One of the largest challenges for conducting tokamak research and future commercialization are plasma \textit{disruptions}, or the loss of plasma stability that may deposit large amounts of temperature and current on the tokamak's walls  \cite{maris2023impact}. Quenches can inflict severe damage to the tokamak, requiring costly repairs and delaying future experiments.  For more background,  please see the Appendix.

Disruptions are difficult to model using physical or ``first-principles'' simulations due to the numerous causes of instability and unobserved factors. In response, several Machine Learning strategies have been previously suggested to estimate \textit{disruptivity}, i.e. the probability of a disruption outcome, allowing timely activation of mitigation systems \cite{cannas2013automatic}. For instance, DIII-D, the U.S.'s largest tokamak, employs a random forest (RF) for plasma state monitoring \cite{reaRealtimeMachineLearningbased2019}. Current state of the art models employ classical neural techniques, such as ``Hybrid Deep Learner'' (HDL), a convolutional Long-Short Term Memory network \cite{zhuHybridDeeplearningArchitecture2020}. These models are still behind what is necessary for commercially viable fusion, where recall of at least 95\% is required \cite{maris2023impact}. Moreover, they might not be tailored to a plasma dataset's unique structure. The RF and HDL treat the problem as a discrete time series, where the observation frequency is fundamental to the underlying system (i.e. like characters or words, which are fundamental units of measure in a sentence.) A tokamak's  features' sampling rates are by no means fundamental, and vary depending on diagnostic instrument. 

The model that we present here, the Continuous CNN \cite{kniggeModellingLongRange2023}, addresses these shortcomings. Instead of learning discrete filters directly, this model learns real, continuous functions that are that sampled. In our case, this is a mapping $\varphi: \mathbb{R} \rightarrow \mathbb{R}$ that maps time to filter weight. Here we use the Multiplicative Anisotropic Gabor Network (MAGNet) \cite{romeroFlexConvContinuousKernel2022}, that uses a combination of 1D convolutions and Gabor functions, (a product of a normal PDF and $\sin$). Learning a MAGNet instead of a discrete filter constrains the kernel to learn smooth features, even though it is later reduced to a discrete series of weights when passing through the sample's resolution. 

\begin{figure}
    \centering
    \includegraphics[width=\linewidth]{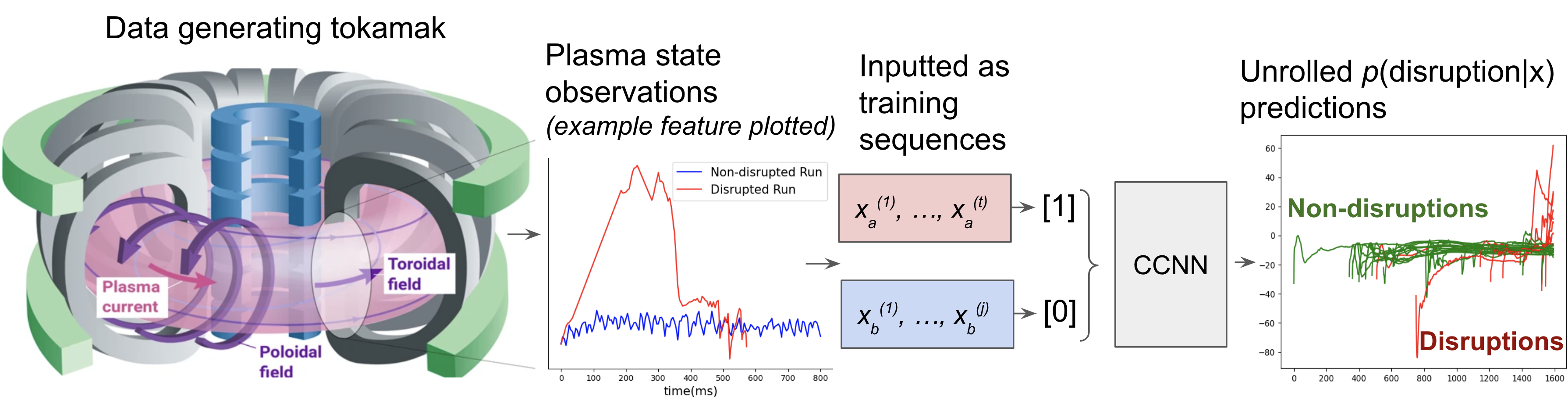}
    \caption{Disruption prediction framework. Tokamak picture taken from \cite{farcacs2022general}.}
    \label{fig:enter-label}
\end{figure}

We endeavor to create a model that surpasses the HDL in the Area Under the Receiver Operating Characteristic Curve (AUC) metric, and also investigate whether a continuous approach is superior. 

\section{Model and Data}

\subsection{Dataset composition}

Our dataset is composed of 4418 plasma shots from the Alcator C-Mod tokamak from MIT (C-Mod) \cite{hutchinsonFirstResultsAlcatorCMOD1994}. Of these, 20\%  culminate in a disruption. The data is normalized to a sample rate of 200Hz (every 5ms). Shots with duration less than 125ms are excluded. Each shot is truncated 40ms prior to its end, as this is the minimum time needed for the activation of disruption mitigation systems. We did not train the model across multiple reactors, as detailed in \cite{zhuHybridDeeplearningArchitecture2020}, due to irregularities noticed in the data other than C-Mod, and because C-Mod is most similar to a new machines of interest \cite{creely2020overview}. Our main objective is to understand how this model performs on a single tokamak, and determine how predictable disruptions actually are when using continuous filters.  
We set up the training and test task as a sequence to label prediction. Each shot is a training example with a binary disruption label. Non-disruptive shots are augmented by randomly clipping them at shorter lengths. For an explanation of the features we use, please see table \ref{tab:feature_tab} in the Appendix. 

\begin{figure}[t]
    \begin{subfigure}[t]{0.55\linewidth}
        \centering
        \includegraphics[width=\linewidth]{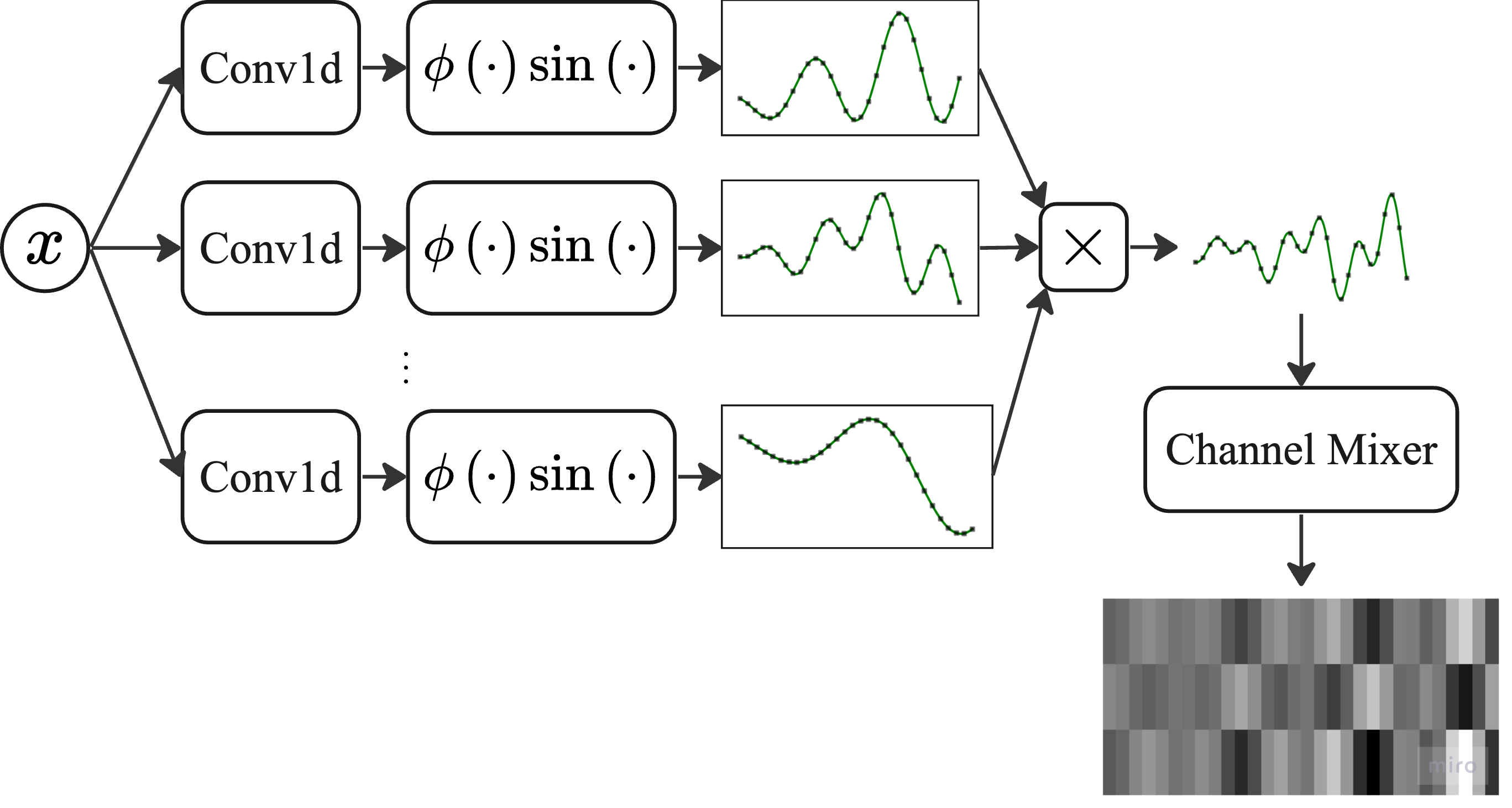}
        \caption{A representation of how the MAGNet kernel is constructed. Coordinates $x$ are transformed into sampled representations of continuous functions, which are then turned into a discrete convolution.}
        \label{fig:magnet}
        \end{subfigure}
        \hspace{.01\textwidth}
    \begin{subfigure}[t]{0.44\linewidth}
        \centering
        \includegraphics[width=0.7\linewidth]{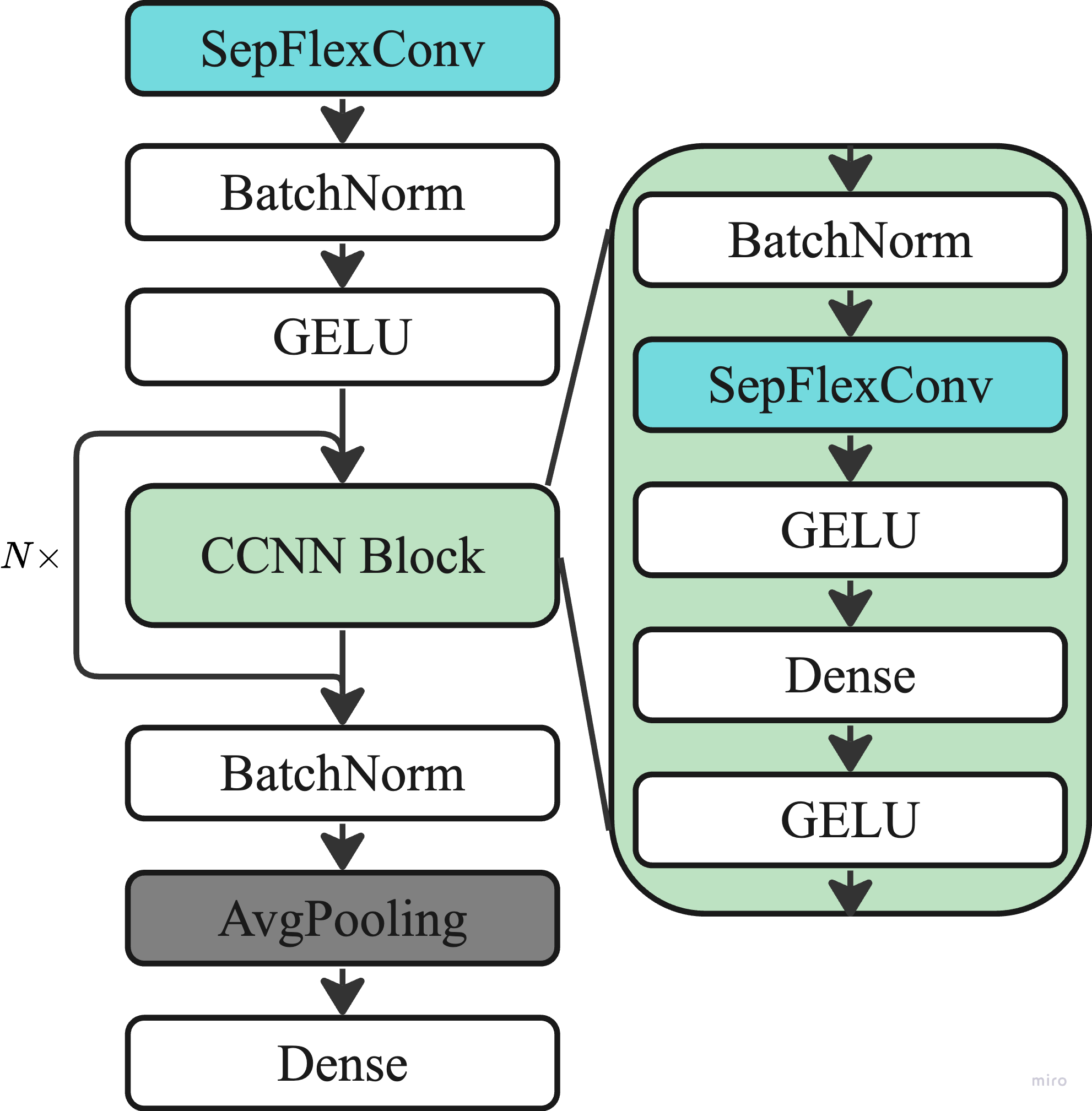}
        \caption{The architecture of a CCNN. Filters are stacked with recurrent connections between dense layers, norms, and nonlinearities.}
        \label{fig:CCNN}
    \end{subfigure}
    \caption{Summary of the CCNN architecture.}
\end{figure}

\subsection{CCNN Architecture}
We instantiated the same architecture as described in \cite{kniggeModellingLongRange2023}. We use separable \texttt{FlexConvs} (\texttt{SepFlexConvs}) from \cite{romeroFlexConvContinuousKernel2022} with MAGNet kernels. A diagram of the MAGNet kernel network is detailed in Fig \ref{fig:magnet}. 
Discretized coordinate values $x$ go through a 1D convolution, and are then passed into the Gabor function $\phi\sin$, where $\phi$ is a normal PDF with learned variance and mean. This is multiplied with the previous filter and passed through another 1D convolution (omitted in the diagram). This repeated $N_f$ times and then passed through a Channel Mixer (another 1D convolution). Another portion of the network computes a continuous mask that truncates the filter at a given threshold. See \cite{romeroFlexConvContinuousKernel2022} for details.
Here, separability refers to the channel mixer inserted at the end of the MAGNet, as show in Fig \ref{fig:magnet}. Non-separable \texttt{FlexConvs} were also explored extensively and no improvement in performance was observed. 

Although the model is trained for sequence-to-label prediction, it can also function as a sequence-to-sequence model. We replace the \texttt{AvgPooling} layer in the CCNN block with a moving average. This approach yielded better results compared to other label representation methods, such as using a vector of 1s or $\tau$-windowing (0 then 1 for $\tau$ timesteps before the end). Modeling problems as sequence-to-label with a \texttt{AvgPooling} layer is observed in multiple high performing global convolution models such as \cite{guEfficientlyModelingLong2022, kniggeModellingLongRange2023}, and is key to performance.

Our network size is far smaller than any of those present in \cite{kniggeModellingLongRange2023}. Our highest performing model has only 2,664 parameters, and increasing model size past did not appear to increase performance. Other than our the final dense layer, nearly our whole model is visualized in Fig \ref{fig:discretizedfilters}. Smaller models are also reasonable performers: a model as small as 994 parameters was found with only a 0.005 drop in AUC. Since the convolution length does not depend on the number of parameters, high performance over long range dependencies is achievable without many variables.
We also found that \cite{kniggeModellingLongRange2023} uses large initialization values for some filters, resulting in sampling artifacts. This may be advantageous common benchmarks like Long Range Arena \cite{tayLongRangeArena2020}, but were very harmful to performance in our model. We lowered the initialization of $\omega_0$ by a factor of 250x and observed increased performance.

\section{Results and Discussion}

\begin{figure}[!htb]
    \begin{subfigure}[T]{0.49\linewidth}
        \centering
        \includegraphics[width=0.6\linewidth]{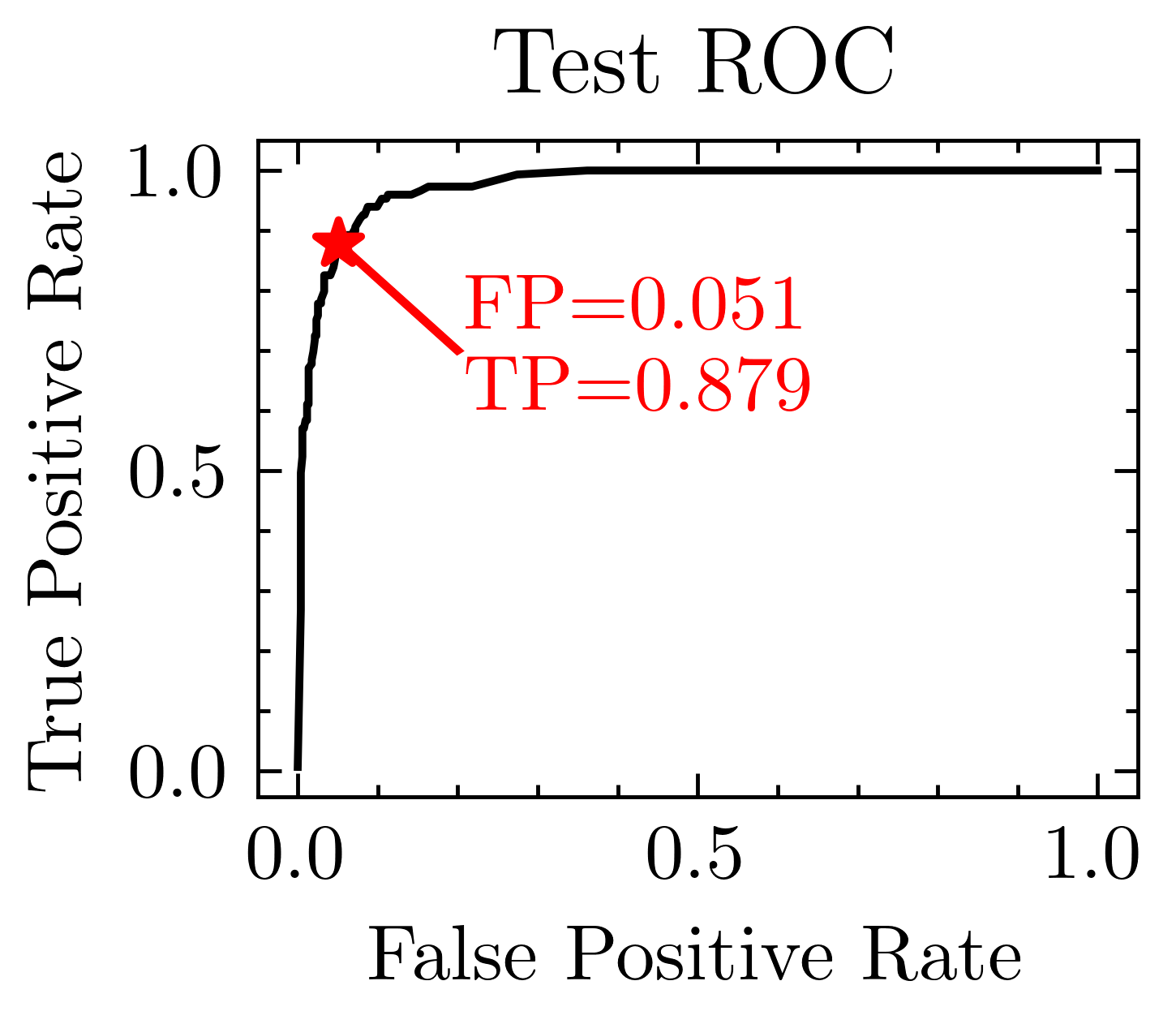}
        \caption{The Test ROC Curve. A recall rate of 87.9\% can be achieved with a false positive rate of 5.1\%}
        \label{fig:roc}
    \end{subfigure}
    \hfill
    \begin{subfigure}[T]{0.49\linewidth}
        \includegraphics[width=\linewidth]{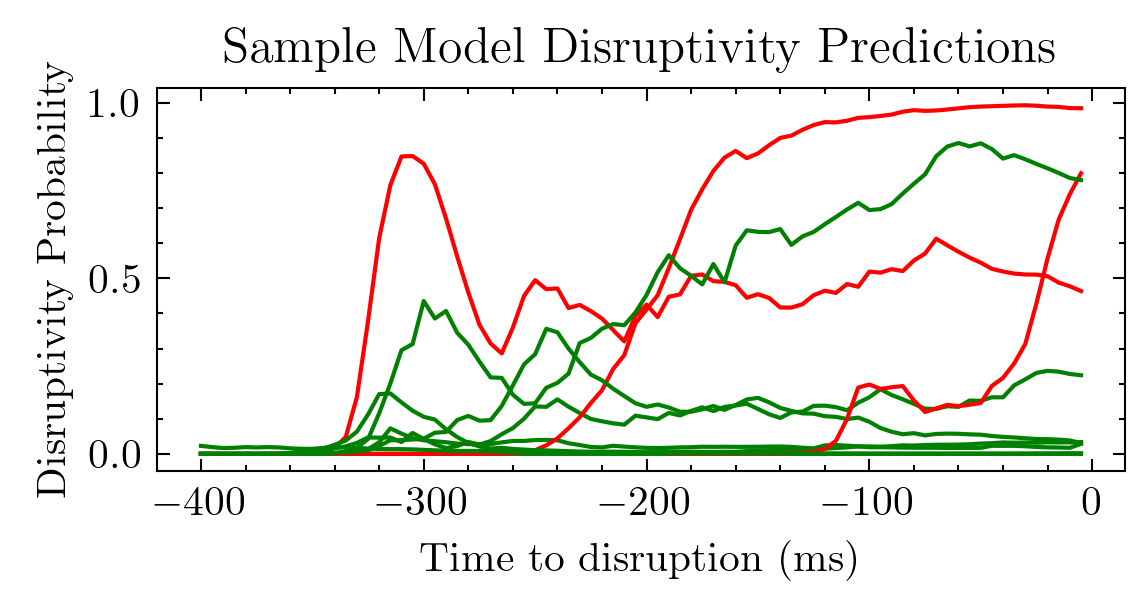}
        \caption{Sample Disruptivity plots over time. $T=0$ is the time of disruption. Red lines denote disruptions, green lines denote non-disruptions.}
        \label{fig:disruptivity}
    \end{subfigure}
    \caption{}
    \label{fig:bigfig}
\end{figure}

Performance is \textit{exceptionally} higher than \cite{zhuHybridDeeplearningArchitecture2020} with an AUC of   \textbf{0.974} on C-Mod compared to 0.799. The ROC curve is shown in Fig \ref{fig:roc}. As indicated, 87.9\% of disruptions can be caught 40ms before disruption with only a 5.1\% false positive rate. As shown in \ref{tab:AUCs}, we perform significantly better than \cite{zhuHybridDeeplearningArchitecture2020} when more disruption data is present. Poor performance on case 2 is likely due to difficulty in dealing with the high class imbalance (\~176 non-disruptions for each disruptive example).

\begin{table}[h]
    \centering
    \begin{tabular}{|c|c|c|c|c|}\hline
    &\multicolumn{2}{c}{Training Set Composition}& \multicolumn{2}{c|}{AUCs} \\\hline
       Case no. & Non-disruptions & Disruptions  &  Baseline \cite{zhuHybridDeeplearningArchitecture2020} & Ours\\\hline
       1 & All& All& .799& \textbf{.974}\\ 
       2 & All& 20& \textbf{.642}& .567\\  
       3 & 33\%& All & .776& \textbf{.915}\\\hline 
       Mean & & & .739& \textbf{.818}\\\hline
    \end{tabular}
    \caption{A table containing different training cases and output metrics}
    \label{tab:AUCs}
\end{table}

The last 400ms of shots are shown in Fig \ref{fig:disruptivity}. Here we see a variety of failure cases that our model encounters. Due to the sequence-to-label nature of our model, it can output positive disruption predictions ($p>0.5$) early in the sequence and reduce this later, as seen in the red bump around $T=-300$ms. In a real-world use, outputting these weights would stop the current shot, and no further data would be observed. Modeling this characteristic of disruption prediction is difficult. Our data augmentation strategy of randomly clipping non-disruptive shots is intended to counteract this, but it is not always successful, especially as we do not augment disruptive examples. We qualitatively observe that most often, the model is able to predict disruptions 100-400ms before they occur. ˙

\subsection{Barriers to real-world use} 
While our model is efficient enough to be used in live disruption prediction (<3ms on CPU for a 1s shot without any optimizations), there remain are barriers to deployment. When generating the dataset, a number of features is obtained by running an equilibrium reconstruction code, i.e. EFIT \cite{laoReconstructionCurrentProfile1985}. EFIT features are available at runtime, but are more prone to generate missing data (represented as white noise) as they rely on more input diagnostics, and we have not yet robustly trained the model on data augmented by the random periods of white noise. Also, there are many data sources that we did not include, including photos and two-dimensional EFIT profiles. These are readily available during experiments, but require significant data-wrangling effort to collect and aggregate. Thus we have set a lower bound the model performance possible in disruption prediction, further motivating the collection of larger, more informative datasets.

\subsection{Continuous vs. Discrete}
\begin{figure}[!htb]
    \centering
    \includegraphics[width=10cm]{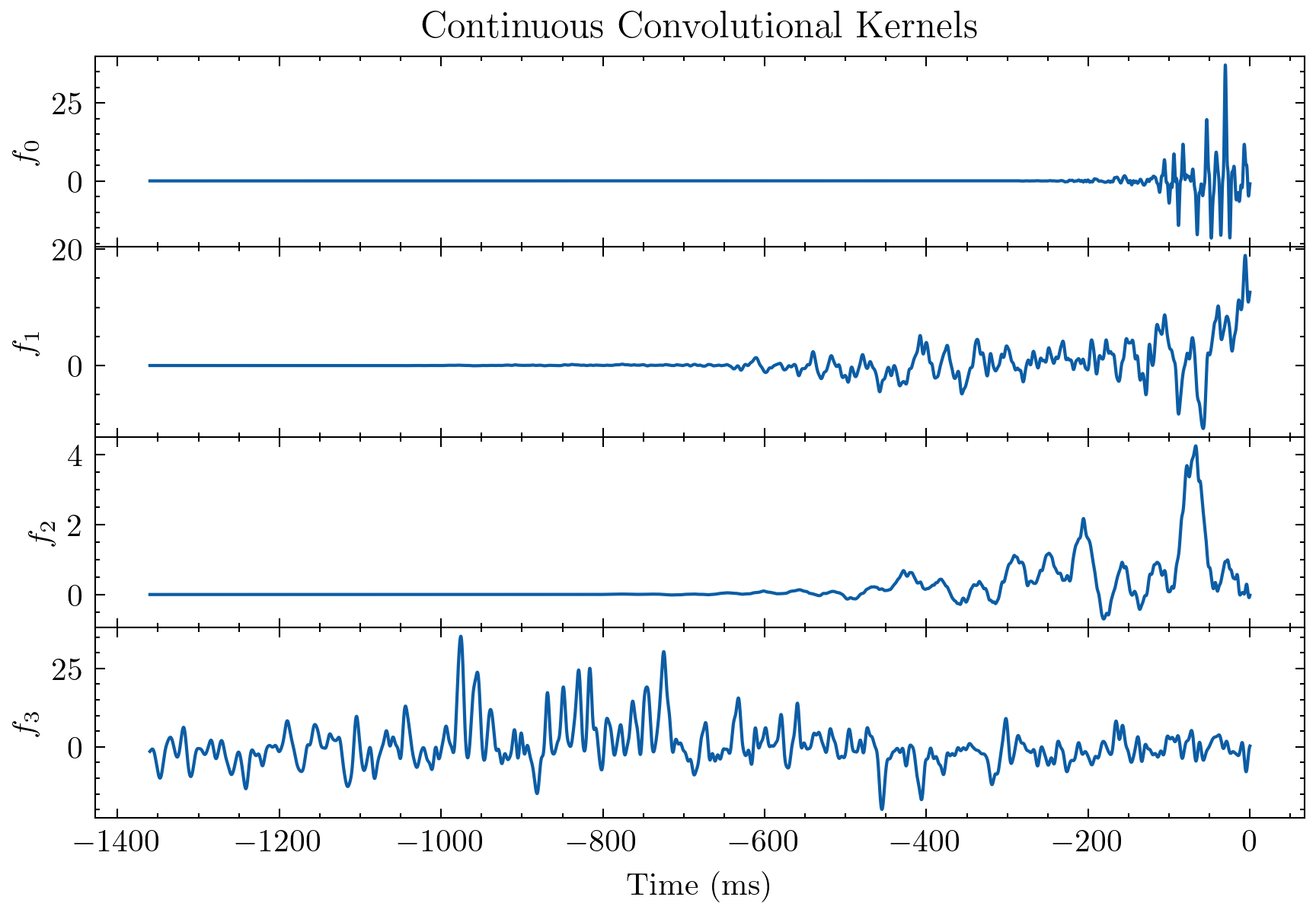}
    \caption{The continuous kernels learned in the order they appear in the model. Notice that $f_0$ is a short, jagged kernel that appears to mimic a derivative. We hypothesize thus that feature $f_0$ refers to the loop voltage feature, which becomes increasingly chaotic near disruptions but is stable for non-disruptions. The higher filters are more abstract combinations of the previous, so they do not map cleanly onto input features. We have generally seen similar behavior from $f_0$ throughout }
    \label{fig:cont-filters}
\end{figure}

\begin{figure}[!htb]
    \centering
    \includegraphics[height=8cm]{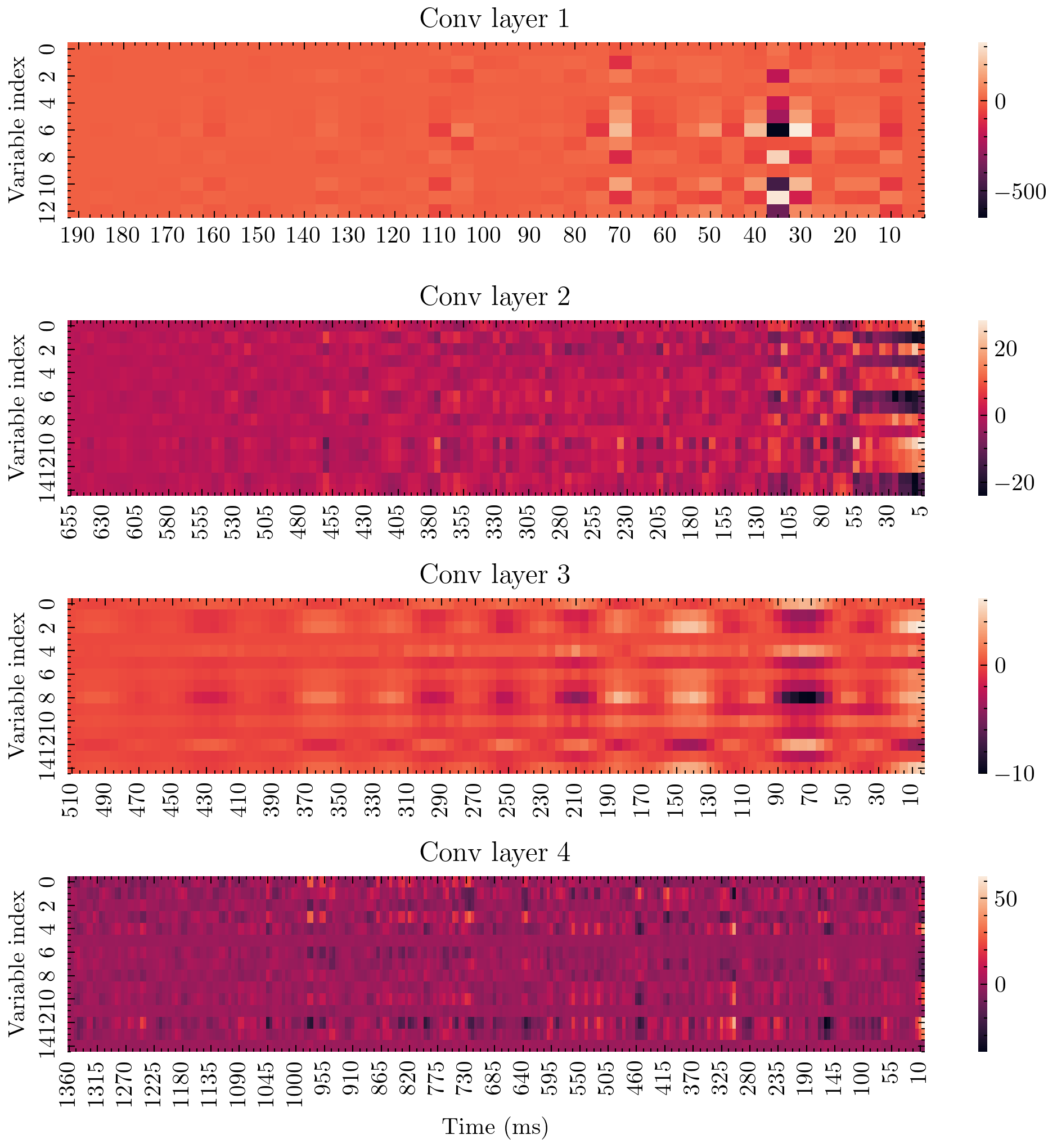}
    \caption{The discretized forms of the kernels. They are quite long: all over 190ms, and most over 500. The first filter takes the form of a  derivative with it's main feature being a single large bump across one timestep. The other filters have much more complicated, long term structure.}
    \label{fig:discretizedfilters}
\end{figure}

Creating filters of arbitrary length with a very small number of parameters is extremely powerful for the task of disruption prediction. Our data size is inherently limited, and training a model with many discrete convolutions is unlikely to generalize well. We learn interpretable, continuous filters (see Fig \ref{fig:cont-filters}), which are significantly longer range than what would be feasible using a discrete models, at 190ms, 655ms, 510ms, and 1360ms long (see Fig \ref{fig:discretizedfilters}). Hence we believe continuous models show a distinct advantage for disruptivity prediction.

\subsection{Architecture limitations}
Many of the samples trained on were quite short (less than a second). Using a moving average before the dense layer at the end of the model for exceedingly long sequences may eventually limit the models ability to signal useful information due to an excessively large denominator. Solutions such as a windowed average, exponential average, or even a simple sum will be explored in the future.

\section{Conclusion} 

Overall, this contribution could support the fight against climate change by bringing fusion energy closer to commercial viability. We present a new state-of-the-art model for disruption prediction based on the CCNN. It achieves greater performance metrics than a baseline, and motivates the development of future models, built with more data, which may greatly reduce harm of disruptions. 

\begin{ack}
We thank PSFC for providing ample compute resources, William Brandon for advice on model setup, Andrew Maris for advice on physics, and Matteo Bonotto, Tommasso Gallingani, Daniele Bigone, and Francesco Cannarile for consistent technical collaboration, advice, ideas and feedback. This work was supported by Eni S.p.A. through the MIT Energy Initiative and by Commonwealth Fusion Systems under SPARC RPP021 funding. This work was supported by Institute of Information \& communications Technology Planning \& Evaluation (IITP) grant funded by the Korea government (MSIT) (No.2023-0-00075, Artificial Intelligence Graduate School Program(KAIST)).
\end{ack}

\printbibliography

@article{creely2020overview,
  title={Overview of the SPARC tokamak},
  author={Creely, AJ and Greenwald, Martin J and Ballinger, Sean B and Brunner, D and Canik, J and Doody, Jeffrey and F{\"u}l{\"o}p, T and Garnier, DT and Granetz, R and Gray, TK and others},
  journal={Journal of Plasma Physics},
  volume={86},
  number={5},
  pages={865860502},
  year={2020},
  publisher={Cambridge University Press}
}

@article{farcacs2022general,
  title={A general framework for quantifying uncertainty at scale},
  author={Farca{\c{s}}, Ionu{\c{t}}-Gabriel and Merlo, Gabriele and Jenko, Frank},
  journal={Communications Engineering},
  volume={1},
  number={1},
  pages={43},
  year={2022},
  publisher={Nature Publishing Group UK London}
}

@article{maris2023impact,
  title={The Impact of Disruptions on the Economics of a Tokamak Power Plant},
  author={Maris, Andrew D and Wang, Allen and Rea, Cristina and Granetz, Robert and Marmar, Earl},
  journal={Fusion Science and Technology},
  pages={1--17},
  year={2023},
  publisher={Taylor \& Francis}
}

@article{reaRealtimeMachineLearningbased2019,
  title = {A Real-Time Machine Learning-Based Disruption Predictor in {{DIII-D}}},
  author = {Rea, C. and Montes, K. J. and Erickson, K. G. and Granetz, R. S. and Tinguely, R. A.},
  date = {2019-07},
  journaltitle = {Nuclear Fusion},
  shortjournal = {Nucl. Fusion},
  volume = {59},
  number = {9},
  pages = {096016},
  publisher = {{IOP Publishing}},
  issn = {0029-5515},
  doi = {10.1088/1741-4326/ab28bf},
  url = {https://dx.doi.org/10.1088/1741-4326/ab28bf},
  urldate = {2023-09-22},
  langid = {english}
}

@article{zhuHybridDeeplearningArchitecture2020,
  title = {Hybrid Deep-Learning Architecture for General Disruption Prediction across Multiple Tokamaks},
  author = {Zhu, J. X. and Rea, C. and Montes, K. and Granetz, R. S. and Sweeney, R. and Tinguely, R. A.},
  year = {2020},
  month = dec,
  journal = {Nuclear Fusion},
  volume = {61},
  number = {2},
  pages = {026007},
  publisher = {{IOP Publishing}},
  issn = {0029-5515},
  doi = {10.1088/1741-4326/abc664},
  urldate = {2023-09-22},
  langid = {english}
}

@article{GIULIANI2023113554,
  title = {Nuclear {{Fusion}} Impact on the Requirements of Power Infrastructure Assets in a Decarbonized Electricity System},
  author = {Giuliani, U. and Grazian, S. and Alotto, P. and Agostini, M. and Bustreo, C. and Zollino, G.},
  year = {2023},
  journal = {Fusion Engineering and Design},
  volume = {192},
  pages = {113554},
  issn = {0920-3796},
  doi = {10.1016/j.fusengdes.2023.113554}
}

@article{spangher2019characterizing,
  title={Characterizing fusion market entry via an agent-based power plant fleet model},
  author={Spangher, Lucas and Vitter, J Scott and Umstattd, Ryan},
  journal={Energy Strategy Reviews},
  volume={26},
  pages={100404},
  year={2019},
  publisher={Elsevier}
}

@article{cannas2013automatic,
  title={Automatic disruption classification based on manifold learning for real-time applications on JET},
  author={Murari, A and Contributors, JET EFDA},
  journal={Nuclear Fusion},
  volume={53},
  number={9},
  pages={093023},
  year={2013},
  publisher={IOP Publishing}
}

@article{kniggeModellingLongRange2023,
  title = {Modelling {{Long Range Dependencies}} in \${{N}}\${{D}}: {{From Task-Specific}} to a {{General Purpose CNN}}},
  shorttitle = {Modelling {{Long Range Dependencies}} in \${{N}}\${{D}}},
  author = {Knigge, David M. and Romero, David W. and Gu, Albert and Gavves, Efstratios and Bekkers, Erik J. and Tomczak, Jakub M. and Hoogendoorn, Mark and Sonke, Jan-Jakob},
  year = {2023},
  publisher = {{arXiv}},
  doi = {10.48550/ARXIV.2301.10540},
  urldate = {2023-09-23},
  copyright = {arXiv.org perpetual, non-exclusive license}
}

@misc{romeroFlexConvContinuousKernel2022,
  title = {{{FlexConv}}: {{Continuous Kernel Convolutions}} with {{Differentiable Kernel Sizes}}},
  shorttitle = {{{FlexConv}}},
  author = {Romero, David W. and Bruintjes, Robert-Jan and Tomczak, Jakub M. and Bekkers, Erik J. and Hoogendoorn, Mark and {van Gemert}, Jan C.},
  year = {2022},
  month = mar,
  number = {arXiv:2110.08059},
  eprint = {2110.08059},
  primaryclass = {cs},
  publisher = {{arXiv}},
  doi = {10.48550/arXiv.2110.08059},
  urldate = {2023-09-29},
  archiveprefix = {arxiv},
}

@misc{guEfficientlyModelingLong2022,
  title = {Efficiently {{Modeling Long Sequences}} with {{Structured State Spaces}}},
  author = {Gu, Albert and Goel, Karan and R{\'e}, Christopher},
  year = {2022},
  month = aug,
  number = {arXiv:2111.00396},
  eprint = {2111.00396},
  primaryclass = {cs},
  publisher = {{arXiv}},
  doi = {10.48550/arXiv.2111.00396},
  urldate = {2023-09-28},
  archiveprefix = {arxiv}
}

@article{tayLongRangeArena2020,
  title = {Long {{Range Arena}}: {{A Benchmark}} for {{Efficient Transformers}}},
  shorttitle = {Long {{Range Arena}}},
  author = {Tay, Yi and Dehghani, Mostafa and Abnar, Samira and Shen, Yikang and Bahri, Dara and Pham, Philip and Rao, J. and Yang, Liu and Ruder, Sebastian and Metzler, Donald},
  year = {2020},
  month = nov,
  journal = {ArXiv},
  urldate = {2023-09-28}}

@article{hutchinsonFirstResultsAlcatorCMOD1994,
  title = {First Results from {{Alcator-C-MOD}}},
  author = {Hutchinson, I.H. and Boivin, R. and Bombarda, F. and Bonoli, P. and Fairfax, S. and Fiore, C. and Goetz, J. and Golovato, S. and Granetz, R. and Greenwald, M. and Horne, S. and Hubbard, A. and Irby, J. and LaBombard, B. and Lipschultz, B. and Marmar, E. and McCracken, G. and Porkolab, M. and Rice, J. and Snipes, J. and Takase, Y. and Terry, J. and Wolfe, S. and Christensen, C. and Garnier, D. and Graf, M. and Hsu, T. and Luke, T. and May, M. and Niemczewski, A. and Tinios, G. and Schachter, J. and Urbahn, J.},
  year = {1994},
  journal = {Physics of Plasmas},
  volume = {1},
  number = {5},
  pages = {1511--1518},
  issn = {1070-664X},
  doi = {10.1063/1.870701},
  langid = {english},
}

@article{laoReconstructionCurrentProfile1985,
  title = {Reconstruction of Current Profile Parameters and Plasma Shapes in Tokamaks},
  author = {Lao, L. L. and John, H. St and Stambaugh, R. D. and Kellman, A. G. and Pfeiffer, W.},
  year = {1985},
  month = nov,
  journal = {Nuclear Fusion},
  volume = {25},
  number = {11},
  pages = {1611},
  issn = {0029-5515},
  doi = {10.1088/0029-5515/25/11/007},
  urldate = {2023-09-29},
  langid = {english}
}

\newpage

\section*{Appendix}

\subsection{Background on Nuclear Fusion}

There are two main techniques to confine fusion plasmas, both of which are currently in experimental / laboratory development. The first, is \textit{inertial confinement}, using highly energetic lasers to compress and heat a solid target of fuel, or fast moving projectiles.  and \textit{magnetic confinement}, which leverages magnetic fields to confine and heat the ionized gas of fuel. While various designs are under exploration for both methods, our group focuses on magnetic confinement using tokamaks. 

Tokamaks are designed to leverage axisymmetric magnetic-fields, and so they maintain stability under specific conditions. Stability is defined as the plasma's maintaining shape, current, and magnetic fields, and most importantly, manageability with respect to external controls. However, tokamak plasmas may become instable due to a heterogenous range of causes. For instance, if most electrons lag in toroidal rotations compared to poloidal ones, it can induce magnetic forces that unevenly pressurize the plasma. If unchecked, such imbalances can escalate, leading to plasma confinement loss, which leads to large temperature and current quenches deposited on the devices' walls. To name a few other types of instabilities, the plasma may experience vertical displacement events, unexpected material residue deposited on the devices' plasma facing components, or kinks in the toroidal oscillations. Thus, a disruption predictor must be general enough to cover a range of behavior. These predictors may be essential in proving commercial viability of the devices\cite{spangher2019characterizing}.

\subsection{Data summary}

\begin{table}[!htb]
    \centering
    \begin{tabular}{|c|c|c|}\hline
        Feature & Definition & Relevant instab. \\ \hline \hline

           Locked mode indicator& Locked mode mag. field normalized to toroidal field & MHD \\  

          Rotating mode indicator & Std. dev. of Mirnov array normalized by toroidal field & MHD \\ 

         $\beta_p$ & Plasma pressure normalized by poloidal magnetic field & MHD \\

         $\ell_i$ & Normalized plasma internal inductance & MHD \\ 
         
         $q_{95}$ & Safety factor at 95th flux surface & MHD \\ \hline

         $n/n_G$ & Electron density normalized by Greenwald density & Dens. limit \\ 

         $\Delta z_{\textrm{center}}$ & Vertical position error of plasma current centroid & Vert. Stab. \\ 

         $\Delta z_{\textrm{lower}}$ & Gap between plasma and lower divertor & Shaping \\  

         $\kappa$ & Plasma elongation & Shaping \\ \hline

         $P_{\textrm{rad}}/P_{\textrm{input}}$ & Radiated power normalized by input power & Impurities \\ 

        $I_{\textrm{p,error}}/I_{\textrm{p,prog}} $ & Plasma current error normalized by programmed current & Impurities \\ 

         $V_{\textrm{loop}}$ & Toroidal ``loop'' voltage & Impurities \\ \hline

    \end{tabular}
    \caption{The input features of the model, their definitions, and a categorization of the type of instability the signal indicates.}
    \label{tab:feature_tab}
\end{table}

\subsection{Learned Filters}


\end{document}